\documentclass[sigconf]{aamas}  % do not change this line!

\AtBeginDocument{%
  \providecommand\BibTeX{{%
    \normalfont B\kern-0.5em{\scshape i\kern-0.25em b}\kern-0.8em\TeX}}}

\usepackage{booktabs}
\usepackage{flushend} % do not change this line!
\usepackage{url}
\usepackage{subfig}
\usepackage{algorithm}               % package for pseudocode
\usepackage[noend]{algpseudocode}    % package for pseudocode
\usepackage{xcolor}
\usepackage{threeparttable}
\usepackage{balance}                 % balances reference page

\setcopyright{ifaamas}  % do not change this line!
\copyrightyear{2020} % do not change this line!
\acmYear{2020} % do not change this line!
\acmDOI{} % do not change this line!
\acmPrice{} % do not change this line!
\acmISBN{} % do not change this line!
\acmConference[AAMAS'20]{Proc.\@ of the 19th International Conference on Autonomous Agents and Multiagent Systems (AAMAS 2020)}{May 9--13, 2020}{Auckland, New Zealand}{B.~An, N.~Yorke-Smith, A.~El~Fallah~Seghrouchni, G.~Sukthankar (eds.)}  % do not change this line!

\begin{document}

\title{Integrating Behavior Cloning and Reinforcement Learning for\\Improved Performance in Dense and Sparse Reward Environments}  % put your title here!
%\titlenote{Produces the permission block, and copyright information}

% AAMAS: as appropriate, uncomment one subtitle line; see camera ready instructions
%\subtitle{Extended Abstract}
%\subtitle{Blue Sky Ideas Track}
%\subtitle{JAAMAS Track}
%\subtitle{Doctoral Consortium}                              
%\subtitle{Demonstration}
%\subtitlenote{Please refrain from using subtitle notes}

% AAMAS: submissions are anonymous for most tracks
%\author{Paper \#1596}  % put your paper number here!

%% example of author block for camera ready version of accepted papers: don't use for anonymous submissions
\author{Vinicius G. Goecks}
\affiliation{%
 \institution{Texas A\&M University\\US Army Research Laboratory}
 \city{College Station} 
 \state{Texas} 
 %\country{United States}
}
\email{vinicius.goecks@tamu.edu}

\author{Gregory M. Gremillion}
\affiliation{%
 \institution{US Army Research Laboratory}
 \city{Adelphi} 
 \state{Maryland} 
 %\country{United States}
}
\email{gregory.m.gremillion.civ@mail.mil}

\author{Vernon J. Lawhern}
\affiliation{%
 \institution{US Army Research Laboratory}
 \city{Aberdeen} 
 \state{Maryland} 
 %\country{United States}
}
\email{vernon.j.lawhern.civ@mail.mil}

\author{John Valasek}
\affiliation{%
 \institution{Texas A\&M University}
 \city{College Station} 
 \state{Texas} 
 %\country{United States}
}
\email{valasek@tamu.edu}

\author{Nicholas R. Waytowich}
\affiliation{%
 \institution{US Army Research Laboratory\\Columbia University}
 \city{Aberdeen} 
 \state{Maryland} 
 %\country{United States}
}
\email{nicholas.r.waytowich.civ@mail.mil}

% The example's default list of authors is too long for headers
\renewcommand{\shortauthors}{V. G. Goecks, G. M. Gremillion, V. J. Lawhern, J. Valasek, and N. R. Waytowich}

\begin{abstract}  % put your abstract here!
This paper investigates how to efficiently transition and update policies, trained initially with demonstrations,  using off-policy actor-critic reinforcement learning. 
It is well-known that techniques based on Learning from Demonstrations, for example behavior cloning, can lead to proficient policies given limited data. 
However, it is currently unclear how to efficiently update that policy using reinforcement learning as these approaches are inherently optimizing different objective functions. 
Previous works have used loss functions, which combine behavior cloning losses with reinforcement learning losses to enable this update.
However, the components of these loss functions are often set anecdotally, and their individual contributions are not well understood. 
In this work, we propose the Cycle-of-Learning (CoL) framework that uses an actor-critic architecture with a loss function that combines behavior cloning and 1-step Q-learning losses with an off-policy pre-training step from human demonstrations.
This enables transition from behavior cloning to reinforcement learning without performance degradation and improves reinforcement learning in terms of overall performance and training time. 
Additionally, we carefully study the composition of these combined losses and their impact on overall policy learning. 
We show that our approach outperforms state-of-the-art techniques for combining behavior cloning and reinforcement learning for both dense and sparse reward scenarios. 
Our results also suggest that directly including the behavior cloning loss on demonstration data helps to ensure stable learning and ground future policy updates.
\end{abstract}

% AAMAS: the ACM CCS are encouraged but optional within AAMAS papers


% selected from AAMAS list
  % put your semicolon-separated keywords here!

\maketitle

%%%%%%%%%%%%%%%%%%%%%%%%%%%%%%%%%%%%%%%%%%%%%%%%%%%%%%%%%%%%%%%%%%%%%%%%%%%%%%%%%%%%%%%%%%%%%%%%%%%%%%%%%
%% start of main body of paper

% ------------------------- %
% INTRODUCTION
% ------------------------- %
\section{Introduction}

%\textcolor{red}{TODO: One of the reviewers suggested to shorten p1-p3 (motivation for imitation learning and problems with RL) and expand on the difficulties of integration imitation learning and RL. Make more explicit that this is intended for deployment on physical systems, and their associated constraints.}

Reinforcement Learning (RL) has yielded many recent successes in solving complex tasks that meet and exceed the capabilities of human counterparts, demonstrated in video game environments \cite{Mnih2015a}, robotic manipulators \cite{andrychowicz2018learning}, and various open-source simulated scenarios \cite{lillicrap2015continuous}.
However, these RL approaches are sample inefficient and slow to converge to this impressive behavior, limited significantly by the need to explore potential strategies through trial and error, which produces initial performance significantly worse than human counterparts.
The resultant behavior that is initially random and slow to reach proficiency is poorly suited for real-world applications such as physically embodied ground and air vehicles, or in scenarios where sufficient capability must be achieved in short time spans.
In such situations, the random exploration of the state space of an untrained agent can result in unsafe behaviors and catastrophic failure of a physical system, potentially resulting in unacceptable damage or downtime. 
Similarly, slow convergence of the agent's performance requires exceedingly many interactions with the environment, which is often prohibitively difficult or infeasible for physical systems that are subject to energy constraints, component failures, and operation in dynamic or adverse environments.
These sample efficiency pitfalls of RL are exacerbated even further when trying to learn in the presence of sparse rewards, often leading to cases where RL can fail to learn entirely. 

% combine this paragraph with the next one
One approach for overcoming these limitations is to utilize demonstrations of desired behavior from a human data source (or potentially some other agent) to initialize the learning agent to a significantly higher level of performance than is yielded by a randomly initialized agent. 
This is often termed Learning from Demonstrations (LfD) \cite{argall2009survey}, which is a subset of imitation learning that seeks to train a policy to imitate the desired behavior of another policy or agent. 
LfD leverages data (in the form of state-action tuples) collected from a demonstrator for supervised learning, and can be used to produce an agent with qualitatively similar behavior in a relatively short training time and with limited data. 
This type of LfD, called Behavior Cloning (BC), attempts to learn a mapping between the state-action pairs contained in the set of demonstrations to mimic the behavior of the demonstrator. 
%BC is a subset of LfD where it is desired to learn a mapping from states to actions that mimics the policy of the demonstrator.
LfD also encompasses other learning modalities such as inverse reinforcement learning (IRL), where it is desired to learn the reward function that the agent, or demonstrator, is optimizing to perform the task \cite{Russell1998}.

Though BC techniques do allow for the relatively rapid learning of behaviors that are comparable to that of the demonstrator, they are limited by the quality and quantity of the demonstrations provided and are only improved by providing additional, high-quality demonstrations. 
In addition, BC is plagued by the distributional drift problem in which a mismatch between the learned policy distribution of states and the distribution of states in the training set can cause errors that propagate over time and lead to catastrophic failures. 
By combining BC with subsequent RL, it is possible to address the drawbacks of either approach, initializing a significantly more capable and safer agent than with random initialization, while also allowing for further self-improvement without needing to collect additional data from a human demonstrator. 
However, it is currently unclear how to effectively update a policy initially trained with BC using RL as these approaches are inherently optimizing different objective functions.
Previous works have used loss functions that combine BC losses with RL losses to enable this update, however, the components of these loss functions are often set anecdotally and their individual contributions are not well understood.

%There are different approaches to LfD that have been proposed in the literature that govern how the demonstrations are used for learning \cite{argall2009survey}. Perhaps the simplest form of LfD is \textit{behavior cloning}, which uses the demonstrations to directly learn a function that maps states to actions without learning or specifying a reward function. Behavior cloning can produce an agent with qualitatively similar behavior in a relatively short training time and with limited data across several domains \cite{bojarski2016end,Nakanishi2004,goecks2018cyber,giusti2015machine,wigness2018robot}. An alternative formulation of LfD is \textit{inverse reinforcement learning} (IRL) which treats the LfD problem as a Markov Decision Process (MDP) where the goal is to learn the reward function that best describes the demonstration trajectories \cite{Ng2000}. \textit{Apprenticeship Learning}-based LfD extends IRL to subsequently learning a policy through reinforcement learning based on the learned reward function \cite{Abbeel2004}. This work focuses specifically on methods to effectively integrate behavior cloning-based LfD and RL as there have been several recent works in this area \cite{hester2018deep,vevcerik2017leveraging,Nair2018ICRA,Rajeswaran-RSS-18}, although methods combining IRL-based LfD with RL have been recently proposed \cite{kang2018policy}. 

In this work, we propose the Cycle-of-Learning (CoL) framework, which uses an actor-critic architecture with a loss function that combines behavior cloning and 1-step Q-learning losses with an off-policy algorithm, and a pre-training step to learn from human demonstrations. 
Unlike previous approaches to combine BC with RL, such as \citeauthor{Rajeswaran-RSS-18} \cite{Rajeswaran-RSS-18}, our approach uses an actor-critic architecture to learn both a policy and value function from the human demonstration data, which we show, speeds up learning. 
Additionally, we perform a detailed component analysis of our method to investigate the individual contributions of pre-training, combined losses, and sampling methods of the demonstration data and their effects on transferring from BC to RL.
To summarize, the main contribution of this work are:
\begin{itemize}
    \item We introduce an actor-critic based method, that combines pre-training as well as combined loss functions to learn both a policy and value function from demonstrations, to enable transition from behavior cloning to reinforcement learning.
    \item We show that our method can transfer from BC to RL without performance degradation while improving upon existing state-of-the-art BC to RL algorithms in terms of overall performance and training time.
    %\item an architecture that enables integrating multiple human interaction modalities, namely, learning from demonstrations, interventions, and evaluations;
    \item We perform a detailed analysis to investigate the contributions of the individual components in our method.
    %and highlight the effects of pre-training, combining loss funcitons and demonstration buffer sampling techniques on the perfomance of BC to RL. 
    %We highlight the contributions of the individual components in our method with a detailed ablation study  which examine the effect on overall policy learning of these combined losses in continuous action space environments with dense and sparse reward functions. 
\end{itemize}

Our results show that our approach outperforms BC, Deep Deterministic Policy Gradients (DDPG), and Demonstration Augmented Policy Gradient (DAPG) in two different application domains for both dense- and sparse-reward settings.
% We show that our CoL method was the only method to produce a viable policy for one of the two environments designed specifically to exhibit a high degree of stochasticity. 
% In addition, we show that in dense-reward settings the performance of DDPGfD suffers significantly due to its inclusion of $n$-step Q-learning loss. 
Our results also suggest that directly including the behavior cloning loss on demonstration data helps to ensure stable learning and ground future policy updates, and that a pre-training step enables the policy to start at a performance level greater than behavior cloning.

% % DONE - Item proposed during rebuttal phase
% \textcolor{green}{[DONE] }\textcolor{blue}{REBUTTAL TEXT: Further, addressing comments by [R3], the authors wanted to briefly clarify the differences between behavior cloning (BC) and LfD. BC is a subset of LfD where it is desired to learn a mapping from states to actions that mimics the policy of the demonstrator. LfD also encompasses other learning modalities as inverse reinforcement learning (IRL), where it is desired to learn the cost function used by the demonstrator to perform the task. We will expand this discussion on Section 1 of the final version of this research paper.}

% ------------------------- %
% PRELIMINARIES
% ------------------------- %
\section{Preliminaries}

We adopt the standard Markov Decision Process (MDP) formulation for sequential decision making \cite{Sutton1998}, which is defined as a tuple $(S, A, R, P, \gamma)$, where $S$ is the set of states, $A$ is the set of actions, $R(s,a)$ is the reward function, $P(s'|s, a)$ is the transition probability function and $\gamma$ is a discount factor. 
At each state $s \in S$, the agent takes an action $a \in A$, receives a reward $R(s,a)$ and arrives at state $s'$ as determined by $P(s'|s, a)$. 
The goal is to learn a behavior policy $\pi$ which maximizes the expected discounted total reward. 
This is formalized by the Q-function, sometimes referred to as the state-action value function:
\begin{equation*}
    Q^{\pi} (s,a) = \mathbb{E}_{a_t\sim\pi}\left[\sum_{t=0}^{+\infty}\gamma^tR(s_t,a_t)\right]
\end{equation*}

\noindent taking the expectation over trajectories obtained by executing the policy $\pi$ starting at $s_0 = s$ and $a_0 = a$.

Here we focus on actor-critic methods which seek to maximize
\begin{equation*}
    J(\theta) = \mathbb{E}_{s\sim\mu}[Q^{\pi(.|\theta)}(s,\pi(s|\theta))]
\end{equation*}

\noindent with respect to parameters $\theta$ and an initial state distribution $\mu$.
The Deep Deterministic Policy Gradient (DDPG) \cite{lillicrap2015continuous} is an off-policy actor-critic reinforcement learning algorithm for continuous action spaces, which calculates the gradient of the Q-function with respect to the action to train the policy. 
DDPG makes use of a replay buffer to store past state-action transitions and target networks to stabilize Q-learning \cite{Mnih2015a}. 
Since DDPG is an off-policy algorithm, it allows for the use of arbitrary data, such as demonstrations from another source, to update the policy. 
A demonstration trajectory is a tuple $(s, a, r, s')$ of state $s$, action $a$, the reward $r = R(s,a)$ and the transition state $s'$ collected from a demonstrator's policy.
In most cases these demonstrations are from a human observer, although in principle these demonstrations can come from any existing agent or policy. 

% ------------------------- %
% RELATED WORK
% ------------------------- %
\section{Related Work}

Several works have shown the efficacy of combining behavior cloning with reinforcement learning across a variety of tasks. 
One of the earliest works in this area was by \citeauthor{Schaal1996} \cite{Schaal1996}, who studied demonstration learning and model-based reinforcement learning and their application to classical tasks such as cart-pole. 
Similarly, \citeauthor{atkeson1997robot} \cite{atkeson1997robot}, in a robotic arm swinging up a pendulum task, trained a dynamics model and reward function from human demonstrations to learn a policy and improve it with reinforcement learning.
\citeauthor{Kim2013} \cite{Kim2013} used expert samples to constrain the approximate policy iteration step and learn a value function, parametrized by linear radial basis functions (RBF), with convex optimization.
Recent work by \citeauthor{hester2018deep} \cite{hester2018deep}, known as Deep Q-learning from Demonstrations (DQfD), combined behavior cloning with deep Q-learning \cite{Mnih2015a} to learn policies for Atari games by leveraging a loss function that combines a large-margin supervised learning loss function, 1-step Q-learning loss, and an $n$-step Q-learning loss function that helps ensure the network satisfies the Bellman equation. 
This work was extended to continuous action spaces by \citeauthor{vevcerik2017leveraging} \cite{vevcerik2017leveraging} with DDPG from Demonstrations (DDPGfD), who proposed an extension of DDPG \cite{lillicrap2015continuous} that uses human demonstrations, and applied their approach to object manipulation tasks for both simulated and real robotic environments. 
The loss functions for these methods include the $n$-step Q-learning loss, which is known to require on-policy data to accurately estimate. 
Similar work by \citeauthor{Nair2018ICRA} \cite{Nair2018ICRA} combined behavior cloning-based demonstration learning, goal-based reinforcement learning, and DDPG for robotic manipulation of objects in a simulated environment.

% % DONE - Item proposed during rebuttal phase
% \textcolor{green}{[DONE]}\textcolor{blue}{REBUTTAL TEXT: Based on comments by [R1], we acknowledge that we may have missed relevant existing approaches and plan to expand the related work section to also include comparison with non-deep learning methods that combines learning from demonstration (LfD) and reinforcement learning (RL) and with adversarial imitation learning methods, as pointed out by [R2].}

The Normalized Actor Critic \cite{Gao2018} uses principles from maximum entropy reinforcement learning \cite{Haarnoja2018} and proposes a learning objective which better normalizes the Q-function learned from demonstration data. In addition they proposed a single unified loss function as opposed to a combined loss function of supervised and reinforcement losses and showed superior performance versus existing works in a Minecraft task and two 3D driving tasks. Policy Optimization with Demonstrations (POfD) \cite{kang2018policy} specifies a demonstration learning approach using an adversarial learning objective, seeking to minimize the difference between the learned policy and the demonstration policy when the reward signal is sparse, an approach similar in nature to Generative Adversarial Imitation Learning (GAIL) \cite{ho2016generative}.

A method that is very similar to ours is the Demonstration Augmented Policy Gradient (DAPG) \cite{Rajeswaran-RSS-18}, a policy-gradient method that uses behavior cloning as a pre-training step together with an augmented loss function with a heuristic weight function that interpolates between the policy gradient loss, computed using the Natural Policy Gradient \cite{Kakade2001}, and behavior cloning loss.
They apply their approach across four different robotic manipulations tasks using a 24 Degree-of-Freedom (DoF) robotic hand in a simulator and show that DAPG outperforms DDPGfD \cite{vevcerik2017leveraging} across all tasks. 
Their work also showed that behavior cloning combined with Natural Policy Gradient performed very similarly to DAPG for three of the four tasks considered, showcasing the importance of using a behavior cloning loss both in pre-training and policy training. 

% % DONE - Item proposed during rebuttal phase
% \textcolor{green}{[DONE]}\textcolor{blue}{REBUTTAL TEXT: As pointed by [R2, R3], we agree that the current comparison with similar methods as, for example, DAPG, does not fully highlight the main differences and the novel aspects of the proposed method, the Cycle-of-Learning (CoL) and that more description is needed. For the final version of this work, we will add more details and clarify how our approach differs from similar methods as such as DAPG and Nair et al. (2018) for continuous, and DQfD and Ape-X DQfD for discrete action-spaces. For now, briefly, we wanted to clarify that our approach was designed for real-world applications with continuous action-spaces where avoiding catastrophic actions is a requirement. The pre-training phase (entailing no policy-driven interaction with the real environment) is an essential phase since it allows the agent to start the reinforcement learning optimization process with at least the behavior cloning performance, which is not the case for Nair et al. (2018) and DAPG. In any case, we acknowledge the contributions of DAPG for combining demonstrations with policy gradient, and Nair et al. (2018) for filtering non-optimal demonstration in the auxiliary losses and adding concepts of HER (Hindsight Experience Replay) to solve sparse-reward tasks.}

In summary, when compared to the main related literature, the Cycle-of-Learning (CoL) algorithm differs from existing algorithms in several ways. First, CoL uses an actor-critic architecture, as opposed to the policy gradient algorithm proposed by DAPG \cite{Rajeswaran-RSS-18}. The actor-critic architecture allows the integration of additional human interaction modalities during training as, for example, evaluative feedback to update the critic and human interventions to update the actor; Second, CoL introduces a pre-training phase the combined loss function and the expert demonstrations are used to train the actor and critic network before interacting with the environment, which is not present on state-of-the-art works such as \citeauthor{Nair2018ICRA} \cite{Nair2018ICRA} and \citeauthor{pohlen2018observe} \cite{pohlen2018observe}. Third, CoL learns in continuous action-space environments as opposed to discrete action-spaces as was done in \citeauthor{hester2018deep} \cite{hester2018deep}, \citeauthor{vevcerik2017leveraging} \cite{vevcerik2017leveraging}, and \citeauthor{pohlen2018observe} \cite{pohlen2018observe}.

\section{Proposed Approach}
The Cycle-of-Learning (CoL) framework is a method for leveraging multiple modalities of human input to improve the training of RL agents. 
These modalities can include human demonstrations, i.e. human-provided exemplar behaviors, human interventions, i.e. interdictions in agent behavior with subsequent partial demonstrations, and human evaluations, i.e. sparse indications of the quality of agent behavior.
These individual mechanisms of human interaction have been previously shown to provide various benefits in learning performance and efficiency \cite{knox2009interactively,macglashan2017interactive,warnell2018deep,goecks2018efficiently,saunders2018trial}. 
The successful integration of these disparate techniques, which would leverage their complementary characteristics, requires a learning architecture that allows for optimization of common objective functions and consistent representations.
An actor-critic framework with a combined loss function, as presented in this work, is such an architecture.

% In this paper, we focus on addressing the known issue of transitioning behavior cloning policies to reinforcement learning by utilizing an actor-critic architecture with a combined BC+RL loss function and pre-training phase for continuous state-action spaces, that can learn in both dense- and sparse-reward environments.
In this paper, we focus on extending the Cycle-of-Learning framework to tackle the known issue of transitioning BC policies to RL by utilizing an actor-critic architecture with a combined BC+RL loss function and pre-training phase for continuous state-action spaces, that can learn in both dense- and sparse-reward environments.
The main advantage of our method is the use of an off-policy, actor-critic architecture to pre-train both a policy and value function, as well as continued re-use of demonstration data during agent training, which reduces the amount of interactions needed between the agent and environment. 
This is an important aspect especially for robotic applications or real-world systems where interactions can be costly.

%\textcolor{orange}{PLACEHOLDER (Feel free to remove, move, or modify): Furthermore, additional modalities of human input can be leveraged to improve learning and mitigate the deficiencies of both BC and RL.
%These can include human intervention, i.e. interdiction of agent behavior and subsequent partial demonstration, and human evaluation, i.e. sparse indication of the quality of agent behavior.
%These mechanisms of human interaction have been be previously shown to provide various benefits in learning performance and efficiency \cite{knox2009interactively,macglashan2017interactive,warnell2018deep,goecks2018efficiently,saunders2018trial}.
%Again, the successful integration of these disparate techniques, which would leverage their complementary characteristics, requires a learning architecture that allows for optimization of common objective functions and consistent representations.
%An actor-critic framework with a combined loss function, as presented in this work, is such an architecture.}

The combined loss function consists of the following components: an expert behavior cloning loss that drives the actor's actions toward previous human trajectories, $1$-step return Q-learning loss to propagate values of human trajectories to previous states, the actor loss, and a $L_2$ regularization loss on the actor and critic to stabilize performance and prevent over-fitting during training. 
The implementation of each loss component and their combination are defined as follows:

\begin{itemize}
    \item \textbf{Expert behavior cloning loss ($\mathcal{L}_{BC}$): } Given expert demonstration subset $\mathcal{D}_E$ of continuous states and actions $s^E$ and $a^E$ visited by the expert during a task demonstration over $T$ time steps
        \begin{align}
        \mathcal{D}_E = \left\{s^E_0, a^E_0, s^E_1, a^E_1, ... , s^E_T, a^E_T\right\},
        \end{align}
        a behavior cloning loss (mean squared error) from demonstration data $\mathcal{L}_{BC}$ can be written as
        \begin{equation} \label{eq:bc_loss}
            \mathcal{L}_{BC}(\theta_\pi) = \frac{1}{2} \left(\pi(s_t|\theta_\pi) - a^E_t) \right)^2
        \end{equation}
        in order to minimize the difference between the actions predicted by the actor network $\pi(s_t)$, parametrized by $\theta_\pi$, and the expert actions $a_{E_t}$ for a given state vector $s_t$.
    
    \item \textbf{$1$-step return Q-learning loss ($\mathcal{L}_1$): }The $1$-step return $R_1$ can be written in terms of the critic network $Q$, parametrized by $\theta_Q$, as
        \begin{equation}
            R_1 = r_t + \gamma Q(s_{t+1},\pi(s_{t+1}|\theta_\pi)|\theta_Q).
        \end{equation}
        In order to satisfy the Bellman equation, we minimize the difference between the predicted Q-value and the observed return from the $1$-step roll-out for a batch of sampled states $\textbf{s}$:
        \begin{equation} \label{eq:1step_loss}
            \mathcal{L}_{Q_{1}}(\theta_Q) = \frac{1}{2} \left( R_1 - Q(\textbf{s}, \pi(\textbf{s}|\theta_\pi)|\theta_Q) \right)^2.
        \end{equation}
    
    \item \textbf{Actor Q-loss ($\mathcal{L}_A$): } It is assumed that the critic function $Q$ is differentiable with respect to the action. Since we want to maximize the Q-values for the current state, the actor loss became the negative of the Q-values predicted by the critic for a batch of sampled states $\textbf{s}$:
        \begin{equation} \label{eq:actor_qloss}
            \mathcal{L}_A(\theta_\pi) = - Q(\textbf{s}, \pi(\textbf{s}|\theta_\pi)|\theta_Q).
        \end{equation}
    
    \item \textbf{$L2$ regularization ($\mathcal{L}_{L2}$): }We also add a L2 regularization term for the actor and critic weights to prevent overfitting and control model complexity:
        \begin{equation} 
            \mathcal{L}_{L2} (\theta_\pi) = \theta_\pi^T \theta_\pi,
        \end{equation}
        \begin{equation} 
            \mathcal{L}_{L2} (\theta_Q) = \theta_Q^T \theta_Q.
        \end{equation}
\end{itemize}

Combining the above loss functions for the Cycle-of-Learning becomes

\begin{align} \label{eq:col_loss}
    \mathcal{L}_{CoL}& (\theta_Q, \theta_\pi) = \lambda_{BC} \mathcal{L}_{BC} (\theta_\pi) + \lambda_A \mathcal{L}_A (\theta_\pi) \nonumber \\
     & +  \lambda_{Q_1} \mathcal{L}_{Q_{1}} (\theta_Q) + \lambda_{L2Q} \mathcal{L}_{L2} (\theta_Q) + \lambda_{L2\pi} \mathcal{L}_{L2} (\theta_\pi).
\end{align}

Our approach starts by collecting contiguous trajectories from expert policies and stores the current and subsequent state-actions pairs, reward received, and task completion signal in a permanent expert memory buffer $\mathcal{D}_E$.
During the pre-training phase, the agent samples a batch of trajectories from the expert memory buffer $\mathcal{D}_E$ containing expert trajectories to perform updates on the actor and critic networks using the same combined loss function (Equations \ref{eq:col_loss}).
This procedure shapes the actor and critic initial distributions to be closer to the expert trajectories and eases the transition from policies learned through expert demonstration to reinforcement learning.

After the pre-training phase, the policy is allowed to roll-out and collect its first on-policy samples, which are stored in a separate first-in-first-out memory buffer with only the agent's samples.
After collecting a given number of on-policy samples, the agent samples a batch of trajectories comprising 25\% of samples from the expert memory buffer and 75\% from the agent's memory buffer.
This fixed ratio guarantees that each gradient update is grounded by expert trajectories.
We opted to use a fixed buffer ratio as in the Ape-X DQfD \cite{pohlen2018observe}, one of the extensions of DQfD \cite{hester2018deep}, which is claimed to be the first RL algorithm to solve the first level of Montezuma Revenge, a challenging ATARI task with sparse rewards.
In our experiments, showed in Section \ref{sec:experiments}, we also compared this fixed buffer ratio with the traditional Prioritized Experience Replay (PER) method and showed that the fixed buffer ratio outperforms PER in the sparse reward scenario.
If a human demonstrator is used, they can intervene at any time the agent is executing their policy, and add this new trajectories to the expert memory buffer.
Samples collected by the agent are added to the agent memory buffer, as usual.

After sampling a batch of trajectories from the expert and agent buffers, we perform model updates using the CoL combined loss. This process is repeated after each interaction with the environment.
The proposed method is summarized in the pseudocode shown in Algorithm \ref{alg:algo}.

% % DONE - Item proposed during rebuttal phase
% \textcolor{green}{[DONE]}\textcolor{blue}{REBUTTAL TEXT: We acknowledge the description of the algorithm can be improved [R1], including a better explanation for the L2 term [R3], which will be done for the final version of the manuscript. With respect to rationality behind using a fixed ratio of human and agent samples in the experience replay [R1, R3], we opted to use a fixed buffer ratio as in the Ape-X DQfD work (Pohlen et al., 2018), one of the extensions of DQfD (Hester et al., 2017), which is claimed to be the first RL algorithm to solve the first level of Montezuma Revenge, a challenging ATARI task with sparse rewards. In our experiments we also compared this fixed buffer ratio with the traditional Prioritized Experience Replay (PER) method and showed that the fixed buffer ratio outperforms PER in the sparse reward scenario.}

%% Cycle-of-Learning Pseudocode
\begin{algorithm}[!tb]
\caption{Cycle-of-Learning (CoL): Transitioning from Demonstrations to Reinforcement Learning}\label{alg:algo}
\begin{algorithmic}[1]

% \Procedure{Main}{}
    \State Input:
        \begin{description}
            \item Environment $env$, training steps $T$, data collection steps $M$, batch size $N$, pre-training steps $L$, CoL hyperparameters $\lambda_{Q_1}$, $\lambda_{BC}$, $\lambda_A$, $\lambda_{L2Q}$, $\lambda_{L2\pi}$, $\tau$, and expert trajectories $\mathcal{D}_E$ (if available).
        \end{description}
    \State Output:
        \begin{description}
            \item Trained actor $\pi(s|\theta_\pi)$ and critic $Q(s,\pi|\theta_Q)$ networks.
        \end{description}
    \State Randomly initialize:
        \begin{description}
            \item Actor network $\pi(s|\theta_\pi)$ and its target $\pi'(s|\theta_{\pi'})$ weights.
            \item Critic network $Q(s,\pi|\theta_Q)$ and its target $Q'(s,\pi'|\theta_{Q'})$ weights.
        \end{description}
    \State Initialize empty agent and expert replay buffers $\mathcal{R}$ and $\mathcal{R}_E$.
    \State Load $\mathcal{R}$ and $\mathcal{R}_E$ with expert trajectories $\mathcal{D}_E$, if available.
    \For {pre-training steps = 1, \dots, $L$}
        \State Call \emph{TrainUpdate}() procedure.
    \EndFor
    \For {training steps = 1, \dots, $T$}
        \State Reset $env$ and receive initial state $s_0$.
        \For {data collection steps = 1, \dots, $M$}
            \State Select action $a_t = \pi(s_t|\theta_\pi)$ according to policy.
            \State Perform action $a_t$ and observe reward $r_t$ and next state $s_{t+1}$.
            \State Store transition $(s_t, a_t, r_t, s_{t+1})$ in $\mathcal{R}$.
            \If {End of episode}
                \State Reset $env$ and receive initial state $s_0$.
            \EndIf
        \EndFor
        \State Call \emph{TrainUpdate}() procedure.
    \EndFor
% \EndProcedure

\Procedure{TrainUpdate()}{}
    \If {Pre-training}
        \State Randomly sample $N$ transitions $(s_i, a_i, r_i, s_{i+1})$ from the expert replay buffer $\mathcal{R}_E$.
    \Else
        \State Randomly sample $N*0.25$ transitions $(s_i, a_i, r_i, s_{i+1})$ from the expert replay buffer $\mathcal{R}_E$ and $N*0.75$ transitions from the agent replay buffer $\mathcal{R}$.
    \EndIf
    \State Compute $\mathcal{L}_{Q_1}(\theta_{Q})$, $\mathcal{L}_{BC}(\theta_{\pi})$, $\mathcal{L}_{A}(\theta_{\pi})$, $\mathcal{L}_{L2}(\theta_{Q})$, $\mathcal{L}_{L2}(\theta_{\pi})$
    \State Update actor and critic networks according to Equation \ref{eq:col_loss}.
    \State Update target networks:
    \begin{align*}
        \theta_{\pi'} \leftarrow \tau\theta_{\pi} + (1-\tau)\theta_{\pi'}, \\
        \theta_{Q'} \leftarrow \tau\theta_{Q} + (1-\tau)\theta_{Q'}.
    \end{align*}
\EndProcedure

\end{algorithmic}
\end{algorithm}

% ------------------------- %
% EXPERIMENTAL RESULTS
% ------------------------- %
\section{Experimental Setup and Results}\label{sec:experiments}

\subsection{Experimental Setup}

% % DONE - Item proposed during rebuttal phase
% \textcolor{green}{[DONE]}\textcolor{blue}{REBUTTAL TEXT: Addressing concerns by [R1] about the similarity of the selected environments, we wanted to clarify that, although the goals of the two tasks are similar, the environments are different in terms of state- and action-spaces and physics (2d vs 3d, with and without wind effects). Additionally, in these environments it was possible to collect human demonstrations of the task. We studied using standard benchmark environments as, for example, MuJoCo and PyBullet locomotion tasks, however, due to the nature of the tasks and number of controls, human demonstrations were not feasible. Further, we are also studying other environments to evaluate the CoL where all these criteria are met.}

% As described in the previous sections, in our approach, we collect contiguous trajectories from expert policies and store them in a permanent memory buffer.
As described in the previous sections, in our approach, the Cycle-of-Learning (CoL), we collect contiguous trajectories from expert policies and store them in a permanent memory buffer.
The policy is allowed to roll-out and is trained with a combined loss from a mix of demonstration and agent data, stored in a separate first-in-first-out buffer.
We validate our approach in three environments with continuous observation- and action-space: Lunar\-Lander\-Continuous-v2 \cite{brockman2016openai} (dense and sparse reward cases) and a custom quadrotor landing task \cite{goecks2018efficiently} implemented using Microsoft AirSim \cite{airsim2017fsr}.

\begin{figure}[!t]
    \centering
    \includegraphics[width=.95\columnwidth]{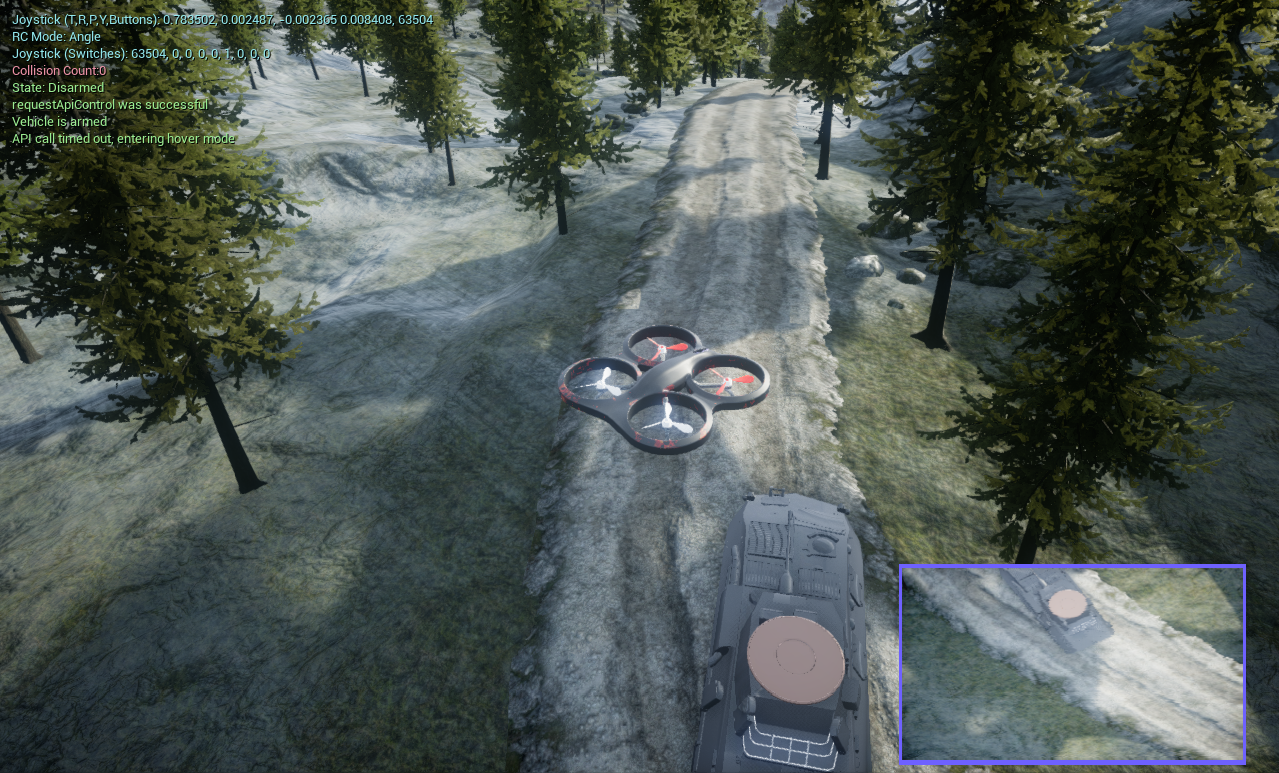}
    \caption{Screenshot of AirSim environment and landing task. Inset image in lower right corner: downward-facing camera view used for extracting the position and radius of the landing pad, which is part of the state space.}
    % \cite{goecks2018efficiently}.}
    \label{fig:Screenshot}
\end{figure}

The dense reward case of Lunar\-Lander\-Continuous-v2 is the standard environment provided by OpenAI Gym library \cite{brockman2016openai}: the state space consists of a eight-dimensional continuous vector with inertial states of the lander, the action space consists of a two-dimensional continuous vector controlling main and side thrusters, and the reward is given at every step based on the relative motion of the lander with respect to the landing pad (bonus reward is given when the landing is completed successfully).
The sparse reward case is a custom modification with the same reward scheme and state-action space, however the reward is stored during the policy roll-out and is only given to the agent when the episode ends and is zero otherwise.
The custom quadrotor landing task is a modified version of the environment proposed by \citeauthor{goecks2018efficiently} \cite{goecks2018efficiently}, implemented using Microsoft AirSim \cite{airsim2017fsr}, which consists of landing a quadrotor on a static landing pad in a simulated gusty environment, as seen in Figure \ref{fig:Screenshot}.
The state space consists of a fifteen-dimensional continuous vector with inertial states of the quadrotor and visual features that represent the landing pad image-frame position and radius as seen by a downward-facing camera.
The action space is a four-dimensional continuous vector that sends velocity commands for throttle, roll, pitch, and yaw.
Wind is modeled as noise applied directly to the actions commanded by the agent and follows a temporal-based, instead of distance-based, discrete wind gust model \cite{moorhouse1980us} with 65\% probability of encountering a wind gust at each time step.
This was done to induce additional stochasticity in the environment. 
The gust duration is uniformly sampled to last between one to three real time seconds and can be imparted in any direction, with maximum velocity of half of what can be commanded by the agent along each axis.
This task has a sparse-reward scheme (reward $R$ is given at the end of the episode, and is zero otherwise) based on the relative distance $r_{rel}$ between the quadrotor and the center of the landing pad at the final time step of the episode: 
\begin{equation*}
    R = \frac{1}{1+r_{rel}^2}.
\end{equation*}
Although the goals of the two tasks are similar, the environments are different in terms of state- and action-spaces and physics (2d vs 3d, with and without wind effects). Additionally, in these environments it was possible to collect human demonstrations of the task. We studied using standard benchmark environments as, for example, MuJoCo and PyBullet locomotion tasks, however, due to the nature of the tasks and number of controls, collecting human demonstrations were not feasible.

% The hyperparameters used in CoL for each environment, and how to tune them properly, are described in Table \ref{tab:exp_hyperparams} in Appendix \ref{ap:hyperparameters}.
% moved supplemental material outside paper due to page limit
The hyperparameters used in CoL for each environment, and how to tune them properly, are described in the project page available online\footnote{Cycle-of-Learning project page: \url{https://vggoecks.com/cycle-of-learning/}.}.

% % DONE - Item proposed during rebuttal phase
% \textcolor{green}{[DONE]}\textcolor{blue}{REBUTTAL TEXT: In terms of hyperparameter tuning [R1, R2], due to sharing similar loss function structure, we started with the same hyperparameters used for a tuned DDPG (Deep Deterministic Policy Gradient) algorithm for the same task and explored the algorithm performance by slightly varying the hyperparameters in the vicinity of those values. Standard hyperparameter tuning routines, like random search, are also valid approaches to tune the CoL for other environments. The lambda parameters [R2], besides standard tuning routines, can be adjusted based on the expertise of the demonstrator (higher LBC for high performing demonstrators, correcting for suboptimal demonstrations [R1]), type of reward function (higher LQ and LA for dense reward schemes), or the level of stochasticity and size of action- and state-spaces (lower LL2 for reduced regularization on smaller and deterministic environments). With respect to how many demonstrations of the task is required [R1], we understand there is no definite method to compute this number as in practice one would use as many as is feasible to collect. By using 20 demonstrations and fewer we wanted to illustrate that the CoL could leverage even a small number of samples.}

\begin{figure*}[!htb]%
    \centering
    \subfloat[]{{\includegraphics[width=0.33\linewidth]{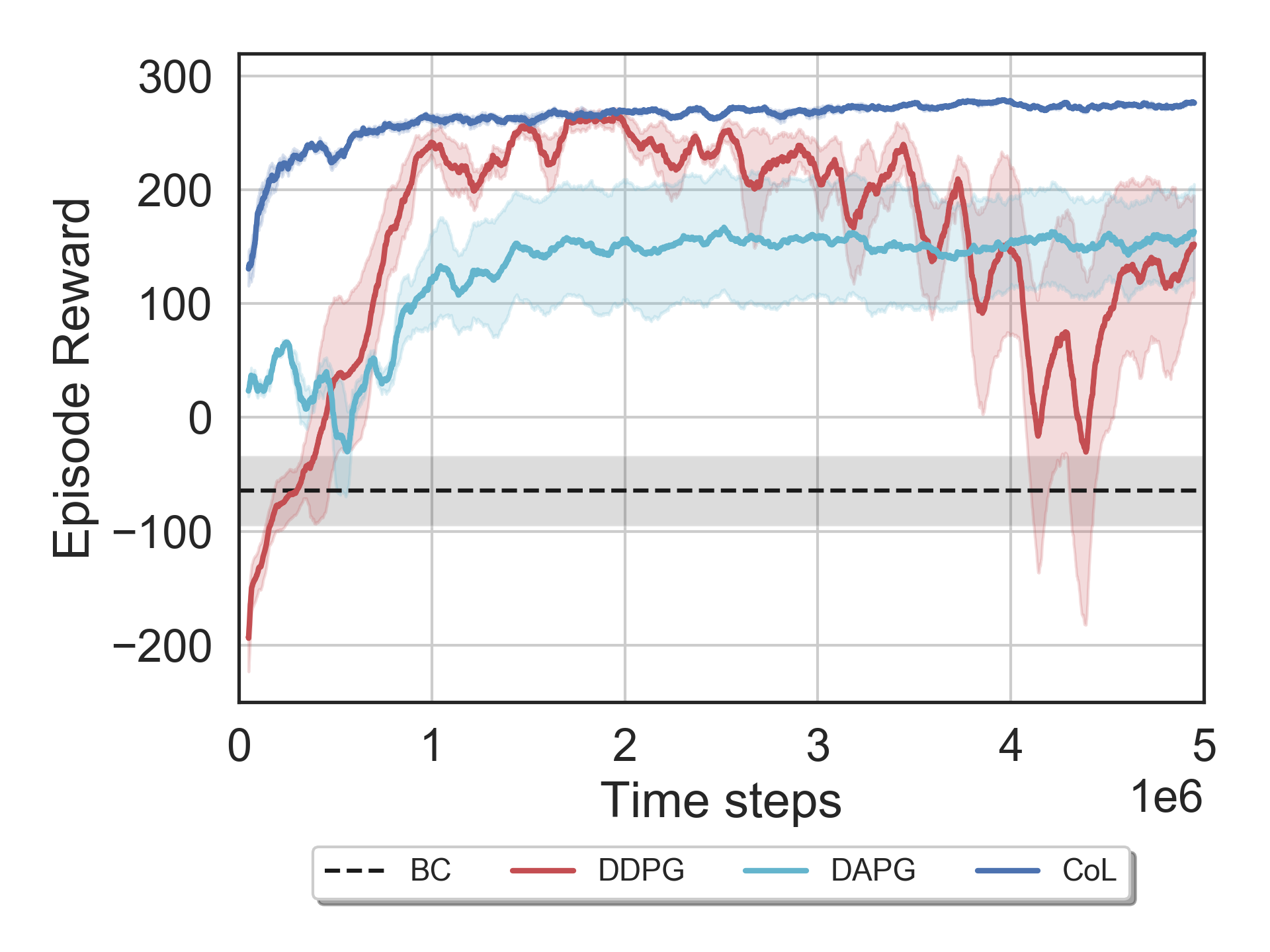}\label{fig:col_results_compact_a} }}%
    \subfloat[]{{\includegraphics[width=0.33\linewidth]{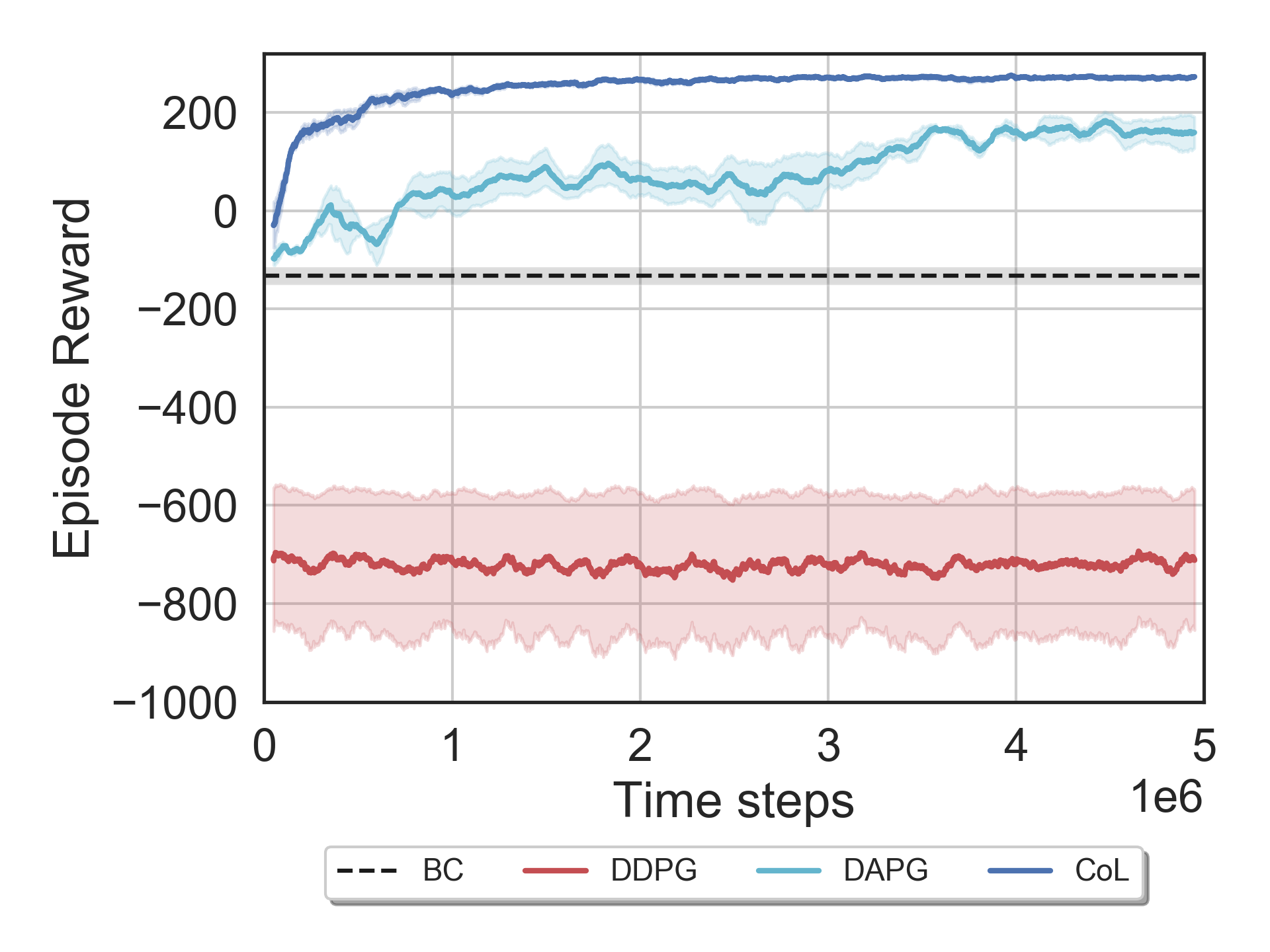}\label{fig:col_results_compact_b} }}%
    \subfloat[]{{\includegraphics[width=0.33\linewidth]{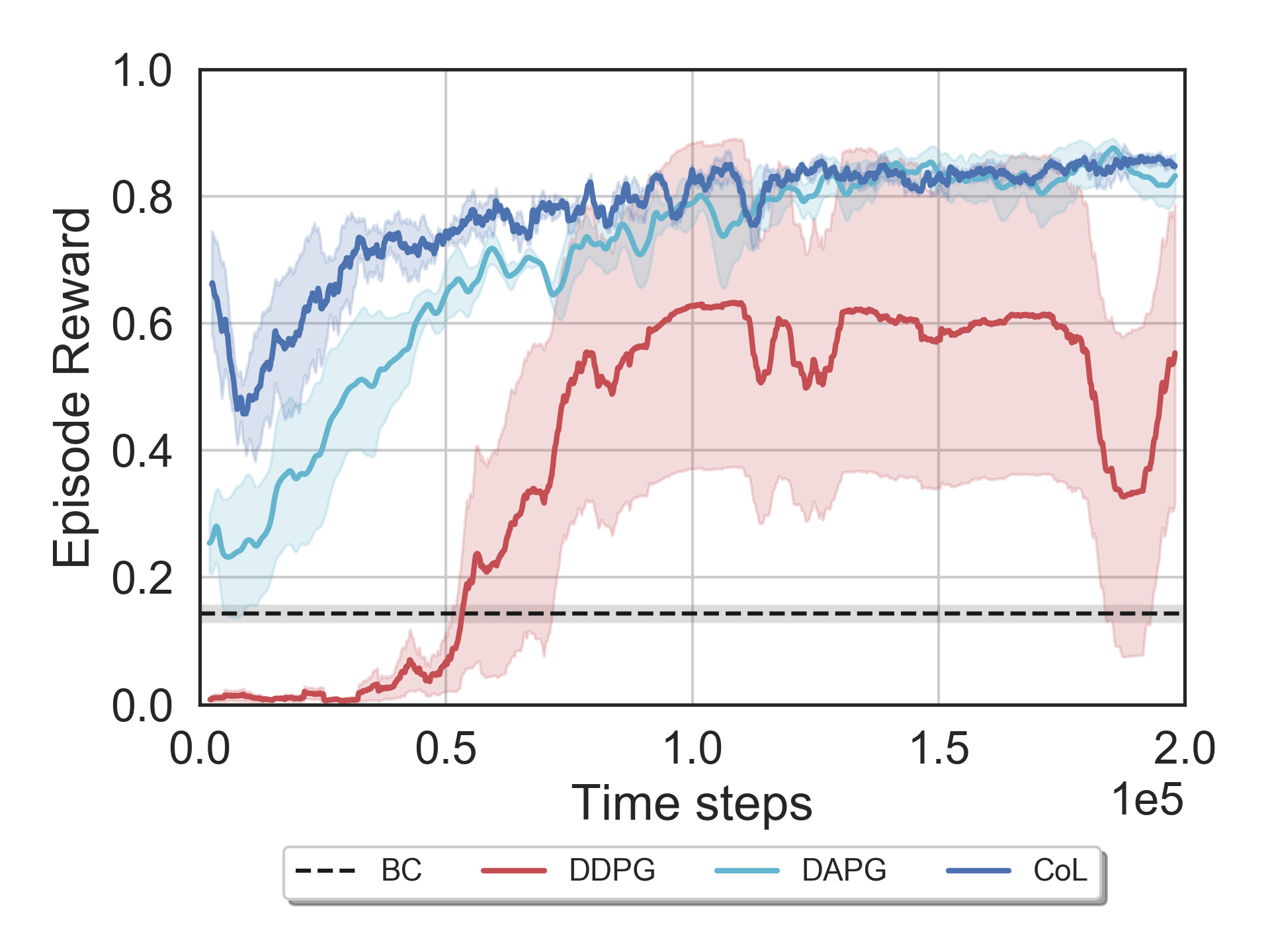}\label{fig:col_results_compact_c} }}%
    \caption{Comparison of CoL, BC, DDPG, and DAPG for 3 random seeds (bold line representing the mean and shaded area the standard error) in the (a) dense- and (b) sparse-reward Lunar\-Lander\-Continuous-v2 environment, and the (c) sparse-reward Microsoft AirSim quadrotor landing environment.}%
    \label{fig:col_results_compact}%
\end{figure*}

The baselines that we compare our approach to are Deep Deterministic Policy Gradient (DDPG) \cite{lillicrap2015continuous,silver2014deterministic}, Demonstration Augmented Policy Gradient (DAPG) \cite{Rajeswaran-RSS-18}, and traditional behavior cloning (BC).
% DDPG
For the DDPG baseline we used an open-source implementation by Stable Baselines \cite{stable-baselines}. 
The hyperparameters used concur with the original DDPG publication \cite{lillicrap2015continuous}: actor and critic networks with 2 hidden layers with 400 and 300 units respectively, optimized using Adam \cite{kingma2014adam} with learning rate of $10^{-4}$ for the actor and $10^{-3}$ for the critic, discount factor of $\gamma = 0.99$, trained with minibatch size of 64, and replay buffer size of $10^{6}$. 
Exploration noise was added to the action following an Ornstein-Uhlenbeck process \cite{uhlenbeck1930theory} with mean of 0.15 and standard deviation of 0.2.
% DAPG
For the DAPG baseline we used an official release of the DAPG codebase from the authors \footnote{DAPG implementation: \url{https://github.com/aravindr93/hand_dapg} \cite{DAPG2018}.}. The policy is represented by a deep neural network with three hidden layers of 128 units each, pre-trained with behavior cloning for 100 epochs, with a batch size of 32 samples, and learning rate of $10^{-3}$, $\lambda_0 = 0.01$, and $\lambda_1 = 0.99$.
% BC
The BC policies are trained by minimizing the mean squared error between the expert demonstrations and the output of the model. 
The policies consist of a fully-connected neural network with 3 hidden layers with 128 units each and exponential linear unit (ELU) activation function \cite{clevert2015fast}. 
The BC policy was evaluated for 100 episodes which was used to calculate the mean and standard error of the performance of the policy.

All baselines that rely on demonstrations, namely BC, DAPG, and CoL, use the same human trajectories collected in the Lunar\-Lander\-Continuous-v2 and custom Microsoft AirSim environment.

\subsection{Experimental Results}

The comparative performances of the CoL against the baseline methods (BC, DDPG and DAPG) for the Lunar\-Lander\-Continuous-v2 environment are presented via their training curves in Figure \ref{fig:col_results_compact_a}, using the standard dense reward.
The mean reward of the BC pre-trained from the human demonstrations is also shown for reference, and its standard error is shown by the shaded band.
The CoL reward initializes to values at or above the BC and steadily improves throughout the reinforcement learning phase.
Conversely, the DDPG RL baseline initially returns rewards lower than the BC and slowly improves until its performance reaches similar levels to the CoL after approximately one million steps.
However, this baseline never performs as consistently as the CoL and eventually begins to diverge, losing much of its performance gains after about four million steps.
The DAPG baseline initial performance, similar to the CoL, surpasses behavior cloning due to the pre-training phase and slowly converges to a high score, although slower than the CoL.

When using sparse rewards, meaning the rewards generated by the Lunar\-Lander\-Continuous-v2 environment are provided only at the last time step of each episode, the performance improvement of the CoL relative to the DDPG and DAPG baselines is even greater (Figure \ref{fig:col_results_compact_b}). 
The performance of the CoL is qualitatively similar during training to that of the dense case, with an initial reward roughly equal to or greater than that of the BC and a consistently increasing reward.
Conversely, the performance of the DDPG baseline is greatly diminished for the sparse reward case, yielding effectively no improvement throughout the whole training period.
The training of the DAPG does not deteriorate when compared to the dense reward case, however, the performance does not match CoL for the specified training time.

The results for the more realistic and challenging AirSim quadrotor landing environment (Figure \ref{fig:col_results_compact_c}) illustrate a similar trend.
The CoL initially returns rewards above the BC, DDPG, and DAPG baselines and steadily increases its performance, with DAPG converging at end to a similar level of performance. 
The DDPG baseline practically never succeeds and subsequently fails to learn a viable policy, while displaying greater variance in performance when compared to CoL and DAPG. 
Noting that successfully landing on the target would generate a sparse episode reward of approximately 0.64, it is clear that these baseline algorithms, with exception of DAPG, rarely generate a satisfactory trajectory for the duration of training.

\subsection{Component Analysis}

\begin{table*}[!htb]
\begin{threeparttable}[t]
\centering
\caption{Method Comparison on Lunar\-Lander\-Continuous-v2 environment, dense-reward case}
\label{tab:ablation_table} 
\begin{tabular}{lllll}\toprule
\bf Method & \bf Pre-Training Loss & \bf Training Loss  & \bf Buffer Type & \bf Average Reward \\\midrule
CoL        & $\mathcal{L}_{Q_{1}} + \mathcal{L}_A + \mathcal{L}_{BC}$     & $\mathcal{L}_{Q_{1}} + \mathcal{L}_A + \mathcal{L}_{BC}$ & Fixed Ratio & 261.80 $\pm$ 22.53  \\
CoL-PT     & None                                                 & $\mathcal{L}_{Q_{1}} + \mathcal{L}_A + \mathcal{L}_{BC}$  & Fixed Ratio & 253.24 $\pm$ 46.50 \\
CoL+PER     & $\mathcal{L}_{Q_{1}} + \mathcal{L}_A + \mathcal{L}_{BC}$     & $\mathcal{L}_{Q_{1}} + \mathcal{L}_A + \mathcal{L}_{BC}$ & PER & 245.24 $\pm$ 37.66 \\
DAPG     & $\mathcal{L}_{BC}$     & Augmented Policy Gradient & None & 127.99 $\pm$ 37.28 \\
DDPG    & None                                    & $\mathcal{L}_{Q_{1}} + \mathcal{L}_A$ & Uniform & 152.98 $\pm$ 69.45  \\
BC         & $\mathcal{L}_{BC}$                                     & None  & None & -48.83 $\pm$ 27.68*  \\
BC+DDPG    & $\mathcal{L}_{BC}$                                     & $\mathcal{L}_{Q_{1}} + \mathcal{L}_A$ & Uniform & -57.38 $\pm$ 50.11  \\
CoL-BC     & $\mathcal{L}_Q{_{1}} + \mathcal{L}_A$                        & $\mathcal{L}_{Q_{1}} + \mathcal{L}_A$  & Fixed Ratio & -105.65 $\pm$ 196.85  \\
\bottomrule
\end{tabular}
\begin{tablenotes}
    \item Summary of learning methods. Enumerated for each method are all non-zero loss components (excluding regularization), buffer type, and average and standard error of the reward throughout training (after pre-training) across the three seeds, evaluated with dense reward in Lunar\-Lander\-Continuous-v2 environment. $^*$For BC, these values are computed from 100 evaluation trajectories of the final pre-trained agent.
\end{tablenotes}
% %\bigskip
% %Should be a caption
\end{threeparttable}%
\end{table*}

Several component analyses were performed to evaluate the impact of each of the critical elements of the CoL on learning.
These respectively include the effects of pre-training, the combined loss function, and the sample composition of the experience replay buffer.
The results of each analysis are shown in Figures \ref{fig:col_pt_cp}-\ref{fig:ablation_results_per} and are summarized in Table \ref{tab:ablation_table}.

\subsubsection{Effects of Pre-Training}
\begin{figure}
    \centering
    \includegraphics[width=0.9\columnwidth]{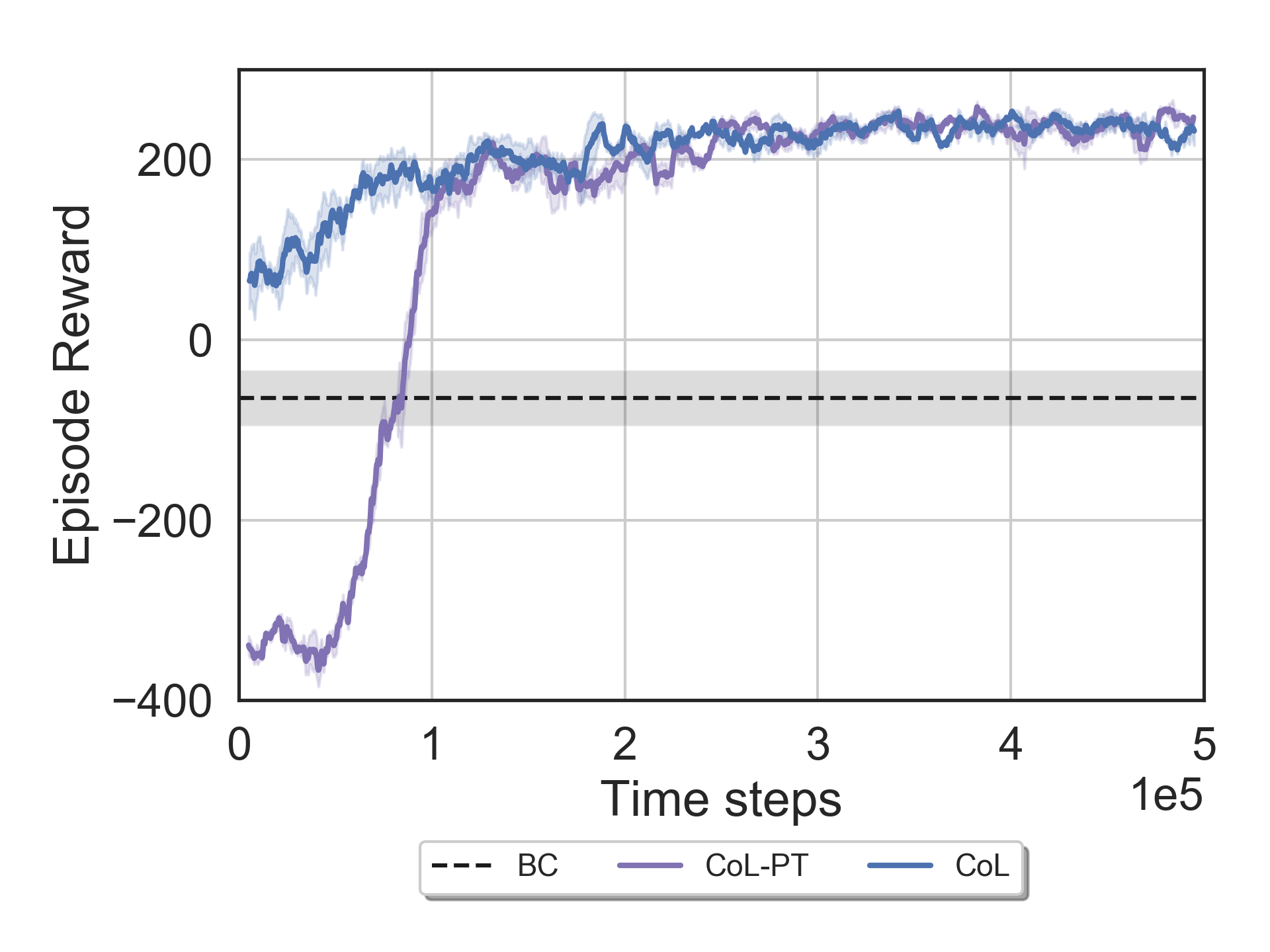}
    \caption{Effects of the pre-training phase in the Cycle-of-Learning. Results for 3 random seeds (bold line representing the mean and shaded area the standard error) showing component analysis in Lunar\-Lander\-Continuous-v2 environment comparing pre-trained Cycle-of-Learning (CoL curve) against the Cycle-of-Learning without the pre-training phase (CoL-PT curve) and the behavior cloning (BC) baseline.}
    \label{fig:col_pt_cp}
\end{figure}

To determine the effects of pre-training on performance we compare the standard CoL against an implementation without this pre-training phase, where the number of pre-training steps $L=0$, denoted as \emph{CoL-PT}. 
The complete combined loss, as seen in Equations \ref{eq:col_loss} is used during the reinforcement learning phase.
This condition assesses the impact on learning performance of not pre-training the agent, while still using the combined loss in the RL phase.
As seen in Figure \ref{fig:col_pt_cp}, this condition differs from the baseline CoL in its initial performance being worse, i.e. significantly below the BC, but does reach similar rewards after several hundred thousand steps, exhibiting the same consistent response during training thereafter.
Effectively, this highlights that the benefit of pre-training is improved initial response and significant speed gain in reaching steady-state performance level, without qualitatively impacting the long-term training behavior.

\subsubsection{Effects of Combined Loss}
\begin{figure}
    \centering
    \includegraphics[width=0.9\columnwidth]{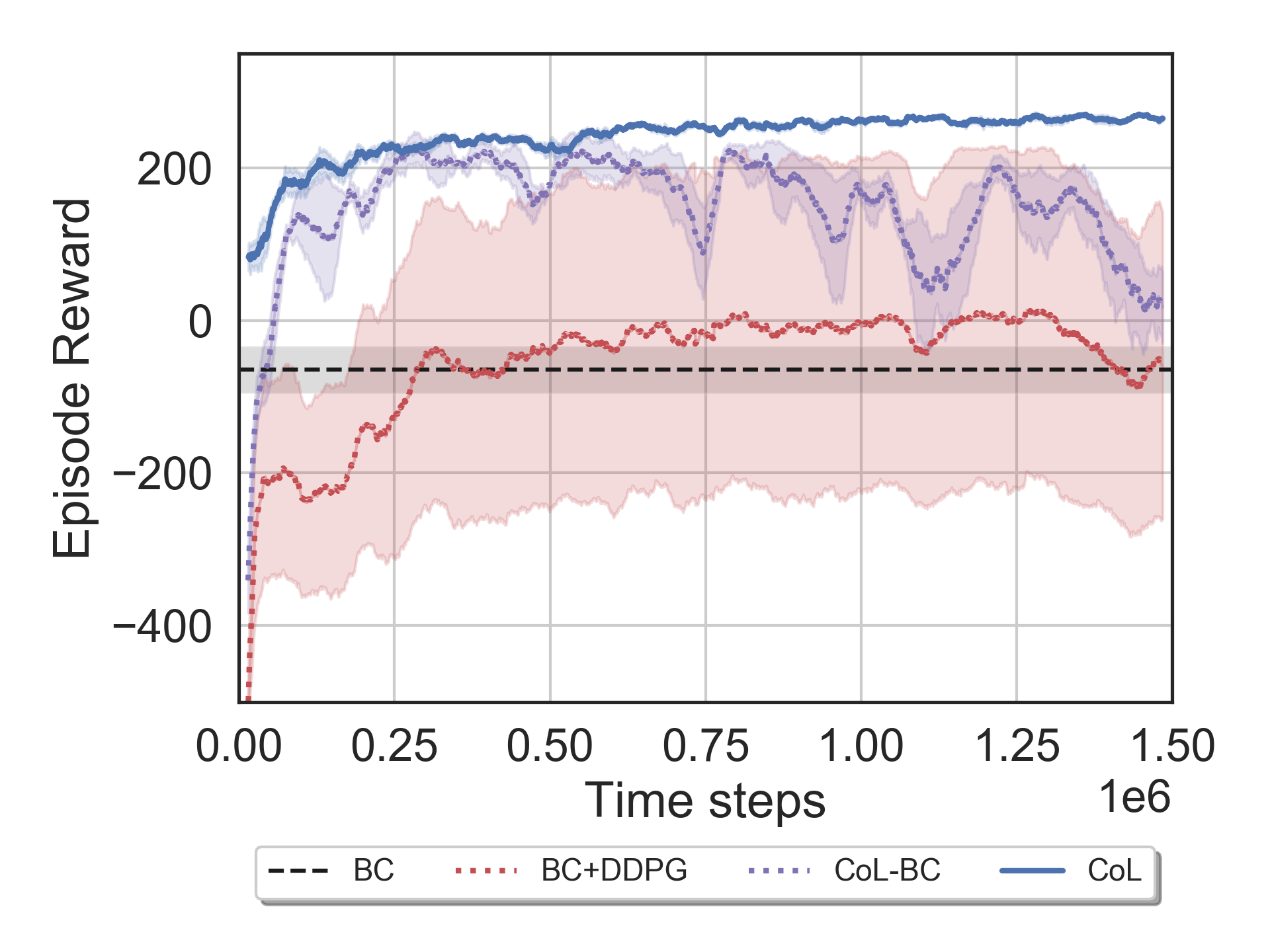}
    \caption{Effects of the combined loss in the Cycle-of-Learning. Results for 3 random seeds (bold line representing the mean and shaded area the standard error) showing component analysis in Lunar\-Lander\-Continuous-v2 environment comparing complete Cycle-of-Learning (CoL), CoL without the expert behavior cloning loss (CoL-BC), and pre-training with BC followed by DDPG without combined loss (BC+DDPG).}
    \label{fig:col_bc_cp}
\end{figure}

To determine the effects of the combined loss function on performance we compare the standard CoL against two alternate learning implementations: 1) the CoL without the behavioral cloning expert loss on the actor ($\lambda_{BC} := 0$) during both pre-training and RL phases, denoted as \emph{CoL-BC}, and 2) standard BC followed by DDPG using standard loss functions, denoted as \emph{BC+DDPG}. 
For the implementation of the CoL without the behavior cloning loss (\emph{CoL-BC}), the critic loss remains the same as in Equation \ref{eq:col_loss} for both training phases.
This condition assesses the impact on learning performance of the behavior cloning loss component $\mathcal{L}_{BC}$, given otherwise consistent loss functions in both pre-training and RL phases.
As seen in Figure \ref{fig:col_bc_cp}, this condition (purple, dashed) improves upon the CoL-PT condition (Figure \ref{fig:col_pt_cp}) in its initial reward return and similarly achieves comparable performance to the baseline CoL in the first few hundred thousand steps, but then steadily deteriorates as training continues, with several catastrophic losses in performance.
This result makes clear that the behavioral cloning loss is an essential component of the combined loss function toward maintaining performance throughout training, anchoring the learning to some previously demonstrated behaviors that are sufficiently proficient.

The second of these comparative implementations that illustrate the effects of the combined loss is the behavior cloning with subsequent DDPG (\emph{BC+DDPG}) condition, which utilized standard loss functions (Equations \ref{eq:bc_loss}, \ref{eq:1step_loss}, and \ref{eq:actor_qloss}) rather than the CoL combined loss in both phases (Equation \ref{eq:col_loss}). 
Pre-training of the actor with BC uses only the regression loss, as seen in Equation \ref{eq:bc_loss}.
%% Commented out DDPG equations to save space. Pointed to their paper instead.
% DDPG utilizes standard loss functions for the actor and critic, as seen in Equation \ref{eq:ddpg_loss}.
DDPG utilizes standard loss functions for the actor and critic, as seen in \citeauthor{lillicrap2015continuous} \cite{lillicrap2015continuous}.
The BC+DDPG condition assesses the impact on learning performance of standardized loss functions rather than our combined loss functions across both training phases.
The BC+DDPG condition (Figure \ref{fig:col_bc_cp}; red, dashed) produces initial rewards below the BC response and subsequently improves in performance only to an average level similar to that of the BC and is much less stable in its response throughout training, as indicated by the wide standard error band. 
This result indicates that simply sequencing standard BC and RL algorithms results in significantly worse performance and stability even after millions of training steps, emphasizing the value of a consistent combined loss function across all training phases.

%% Commented out DDPG equations to save space. Pointed to their paper instead.
% \begin{align} \label{eq:ddpg_loss}
%     \mathcal{L}_{DDPG} (\theta_Q,\theta_\pi) =& \lambda_{Q_1}\mathcal{L}_{Q_1}(\theta_Q) + \lambda_{A}\mathcal{L}_{A}(\theta_\pi)\nonumber \\ 
%     &+ \lambda_{L2Q}\mathcal{L}_{L2}(\theta_Q)+ \lambda_{L2\pi}\mathcal{L}_{L2}(\theta_\pi).
% \end{align}

\subsubsection{Effects of Human Experience Replay Sampling}
\begin{figure}[!t]%
    \centering
    \includegraphics[width=0.9\linewidth]{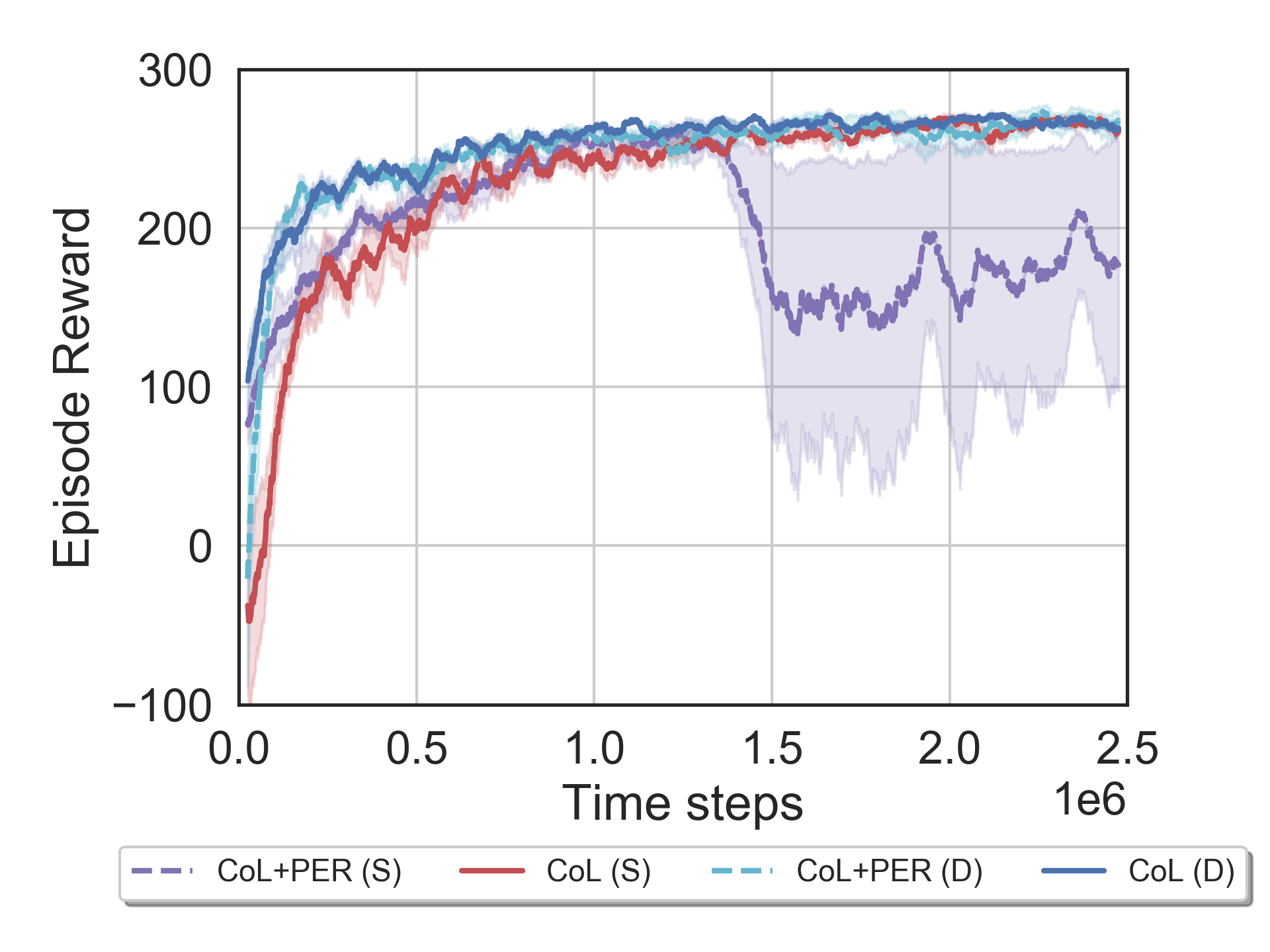}
    \caption{Effects of human experience replay sampling in the Cycle-of-Learning. Results for 3 random seeds (bold line representing the mean and shaded area the standard error) showing ablation study in Lunar\-Lander\-Continuous-v2 environment, dense (D) and sparse (S) reward cases, comparing complete Cycle-of-Learning (CoL) trained with fixed ratio of expert and agent samples and complete Cycle-of-Learning using Prioritized Experience Replay (CoL+PER) with a variable ratio of expert and agent samples ranked based on their temporal difference error.}%
    \label{fig:ablation_results_per}%
\end{figure}

To determine the effects of the different sampling techniques of the experience replay buffer on performance we compare the standard CoL, which utilizes a fixed ratio buffer of samples comprising 25\% expert data and 75\% agent data, against an implementation with Prioritized Experience Replay (PER) \cite{schaul2015prioritized}, with a data buffer prioritized by the magnitude of each transition's temporal difference (TD) error, denoted as \emph{CoL+PER}.
The comparative performance of these implementations, for both the dense- (D) and sparse-reward (S) cases of the Lunar\-Lander\-Continuous-v2 scenario, are shown in Figure \ref{fig:ablation_results_per}.
For the dense-reward condition, there is no significant difference in the learning performance between the fixed ratio and PER buffers.
However, for the sparse-reward case of the \emph{CoL+PER} implementation, the learning breaks down after approximately 1.3 million training steps, resulting in a significantly decreased performance thereafter.
This result illustrates that the fixed sampling ratio for the replay buffer in the standard CoL is a more robust mechanism of incorporating experience data, particularly in sparse-reward environments, likely because it grounds performance to demonstrated human behavior throughout training.

% \begin{equation} \label{eq:col-bc}
%     % \mathcal{L}_{CoL-BC}^A (s) = - Q(s, \pi(s|\theta_\pi)|\theta_Q) + \lambda_{Areg} \mathcal{L}_{reg}^A (\theta_\pi).
%     \lambda_{BC} := 0
% \end{equation}

% \begin{equation} \label{eq:bc_ddpg}
%     \mathcal{L}_{BC}^A (\theta_\pi) = \frac{1}{2} \left(\pi(s|\theta_\pi) - a_{E_t}) \right)^2.
% \end{equation}

% \begin{equation} \label{eq:ddpg_critic}
%     \mathcal{L}_{DDPG}^C (\theta_Q) = \frac{1}{2} \left( R_1 - Q(s, \pi(s|\theta_\pi)|\theta_Q) \right)^2.
% \end{equation}
% \begin{equation} \label{eq:ddpg_actor}
%     \mathcal{L}_{DDPG}^A (\theta_\pi) = - Q(s, \pi(s|\theta_\pi)|\theta_Q).
% \end{equation}

% ------------------------- %
% DISCUSSION AND CONCLUSION
% ------------------------- %
% I think the discussion is the main thing that needs to be worked on at this point. :)
\section{Discussion and Conclusion}
%%% Discussion outline %%%%
%% 1st paragraph: Overview of what we did (i.e. rewording of last paragraph of introduction).
In this work, we present a novel method for combining behavior cloning with reinforcement learning using an actor-critic architecture that implements a combined loss function and a demonstration-based pre-training phase.
%% 2nd paragraph: We compare against state-of-the-art RL and LfD+RL baselines
We compare our approach against state-of-the-art baselines, including BC, DDPG, and DAPG, and demonstrate the superiority of our method in terms of learning speed, stability, and performance with respect to these baselines.
%% 3rd paragraph: We show LfD + RL learning both dense and sparse environmennts
This is shown in the OpenAI Gym Lunar\-Lander\-Continuous-v2 and the high-fidelity Microsoft AirSim quadrotor simulation environments in both dense and sparse reward settings.
%This result holds in both dense and sparse reward settings.% , though the improvements of our method over these baselines is even more dramatic in the sparse case.
This result is especially noticeable in the AirSim landing task (Figure \ref{fig:col_results_compact_c}), an environment designed to exhibit a high degree of stochasticity. 
The BC and DDPG baselines fail to converge to an effective and stable policy after five million training steps on the Lunar\-Lander\-Continuous-v2 environment with dense reward and the modified version with a sparse reward signal.
DAPG, although successful in both Lunar\-Lander\-Continuous-v2 environments and the custom AirSim landing task, converges at a slower rate when compared to the proposed method and starts the training at a lower performance value after pre-training with demonstration data.
Conversely, our method, CoL, is able to quickly achieve high performance without degradation, surpassing both behavior cloning and reinforcement learning algorithms alone, in both dense and sparse reward cases.
%% 4th paragraph We show that both pre-training and the combined loss are necessary to produce these gains
Additionally, we demonstrate through separate analyses of several components of our architecture that pre-training, the use of a combined loss function, and a fixed ratio of human-generated experience are critical to the performance improvements.
This component analysis also indicated that simply sequencing standard behavior cloning and reinforcement learning algorithms does not produce these gains and highlighted the importance of grounding the training to the demonstrated data by using a fixed ratio of expert and agent trajectories in the experience replay buffer.
% %% 5th paragraph: the n-step loss significantly degrades performance. 
% We also illustrate that the lack of a $n$-step Q-learning loss in our architecture is necessary for these improvements.
% Furthermore, we show that inclusion of such a loss term in our algorithm significantly reduces performance in both dense- and sparse-reward conditions, while its omission from DDPGfD significantly improves performance in dense-reward but not in sparse-reward conditions.
% To the best of our knowledge this is the first work that examined the effect of the $n$-step Q-learning loss on learning for DDPGfD policy performance. 

\subsection{Future Work}

Future work will investigate how to effectively integrate multiple forms of human feedback into an efficient human-in-the-loop RL system capable of rapidly adapting autonomous systems in dynamically changing environments. 
Actor-critic methods, such as the CoL method proposed in this paper, provide an interesting opportunity to integrate different human feedback modalities as additional learning signals at different stages of policy learning \cite{waytowich2018cycle}. 
For example, existing works have shown the utility of leveraging human interventions \cite{goecks2018efficiently,saunders2018trial}, and specifically learning a predictive model of what actions to ignore at every time step \cite{Zahavy2018}, which could be used to improve the quality of the actor's policy. 
Deep reinforcement learning with human evaluative feedback has also been shown to quickly train policies across a variety of domains \cite{warnell2018deep,macglashan2017interactive} and can be a particularly useful approach when the human is unable to provide a demonstration of desired behavior but can articulate when desired behavior is achieved. 
% Feedback of this type can be interpreted as a critique of an agent's current behavior relative to the human's expectation of desired behavior \cite{macglashan2017interactive}, thus making it conceptually similar to an advantage function which can be used to improve the quality of the critic. 
Further, the capability our approach provides, transitioning from a limited number of human demonstrations to a baseline behavior cloning agent and subsequent improvement through reinforcement learning without significant losses in performance, is largely motivated by the goal of human-in-the-loop learning on physical robotic systems.
Thus, our aim is to integrate this method onto such systems and demonstrate rapid, safe, and stable learning from limited human interaction.
%% Limitations and Future work?
% - still need to be able to provide demonstrations (input mechanism or existing policy, i.e. Mujoco environments where there are too many DoFs to control)
% - initial performance is still limited by the quantity and quality of demonstrations (possible connection to other CoL stages: intervention and evaluation learning)
% - integration with LfE (and how the loss functions might need to change)
% - implementation in physical system (demonstrate safe transition to RL and further learning)

% ------------------------- %
% ACKNOWLEDGMENTS
% ------------------------- %
\begin{acks}
Research was sponsored by the U.S. Army Research Laboratory and was accomplished under Cooperative Agreement Number W911NF-18-2-0134. The views and conclusions contained in this document are those of the authors and should not be interpreted as representing the official policies, either expressed or implied, of the Army Research Laboratory or the U.S. Government. The U.S. Government is authorized to reproduce and distribute reprints for Government purposes notwithstanding any copyright notation herein.
\end{acks}

%%%%%%%%%%%%%%%%%%%%%%%%%%%%%%%%%%%%%%%%%%%%%%%%%%%%%%%%%%%%%%%%%%%%%%%%%%%%%%%%%%%%%%%%%%%%%%%%%%%%%%%%%
%% bibliography: see CFP for number of permitted pages

\bibliographystyle{ACM-Reference-Format}  % do not change this line!
\balance
\bibliography{sample-bibliography}  % put name of your .bib file here

\end{document}